\documentclass[letterpaper, 10 pt, conference]{ieeeconf}  

\IEEEoverridecommandlockouts  

\usepackage{algorithmic}
\usepackage{amsmath}
\usepackage{amssymb}
\usepackage{bm}
\usepackage{booktabs}
\usepackage[caption=false]{subfig}
\usepackage{color}
\usepackage{graphicx}
\usepackage{tabularx}
\usepackage{tikz}

\DeclareMathOperator*{\argmin}{arg\,min}
\newcommand{\diag}{\mathop{\mathrm{diag}}}

\newcommand\copyrighttext{%
  \footnotesize \textcopyright 2020 IEEE. Personal use of this material is permitted.
  Permission from IEEE must be obtained for all other uses, in any current or future
  media, including reprinting/republishing this material for advertising or promotional
  purposes, creating new collective works, for resale or redistribution to servers or
  lists, or reuse of any copyrighted component of this work in other works.}
  
\newcommand\copyrightnotice{%
\begin{tikzpicture}[remember picture,overlay]
\node[anchor=south,yshift=10pt] at (current page.south) {\fbox{\parbox{\dimexpr\textwidth-\fboxsep-\fboxrule\relax}{\copyrighttext}}};
\end{tikzpicture}%
}

\title{\LARGE \bf
A memory of motion for visual predictive control tasks
}

\author{
Antonio Paolillo$^{1,2}$, Teguh Santoso Lembono$^{1}$ and Sylvain Calinon$^{1}$
\thanks{$^1$Idiap Research Institute, Martigny, Switzerland. $^2$Dalle Molle Institute for Artificial Intelligence (IDSIA), USI-SUPSI, Lugano, Switzerland. E-mail: \texttt{\{name.surname\}@idiap.ch}.} 
\thanks{This work was supported by the European Commission's  H2020 
Programme (MEMMO project, http://www.memmo-project.eu/, grant 780684).}
}

\begin{document}

\maketitle

\copyrightnotice

\thispagestyle{empty}
\pagestyle{empty}

\begin{abstract}
This paper addresses the problem of efficiently achieving visual predictive control tasks. To this end, a memory of motion, containing a set of trajectories built off-line, is used for leveraging precomputation and dealing with difficult visual tasks. Standard regression techniques, such as k-nearest neighbors and Gaussian process regression, are used to query the memory and provide on-line a warm-start and a way point to the control optimization process. The proposed technique allows the control scheme to achieve high performance 
and, at the same time, keep the computational time limited. Simulation and experimental results, carried out with a 7-axis manipulator, show the effectiveness of the approach.
\end{abstract}

\section{INTRODUCTION}\label{sec:intro}
Image-based visual servoing (VS) is a well established technique to control robots using visual information~\cite{Chaumette:ram:06}~\cite{Chaumette:ram:07}.
Its classic formulation consists in the simple control law
%
$\bm{v}=-\lambda \hat{\bm{L}}^\dagger(\bm{s} - \bm{s}^\ast)$,
%
where $\bm{v}$ is the velocity of the camera, $\lambda$ is the control gain and $\bm{L}^\dagger$ is the pseudo-inverse of the image Jacobian (or interaction matrix) $\bm{L}$; the hat symbol denotes an approximation. 
%
%
This control law ensures an exponential convergence to zero of the visual error, i.e., the difference between the measured and desired visual features ($\bm{s}$ and $\bm{s}^\ast$, respectively).
Although the VS control law is easy to implement and fast to execute, it has some limitations.
For large values of the error, the behavior can be unstable, and for some configurations the Jacobian can become singular causing dangerous commands~\cite{Chaumette:chapter:98}.
Being purely reactive, VS does not perform any sort of anticipatory behavior that would improve the tracking performance.
Furthermore, it cannot easily include (visual or Cartesian) constraints, which are very useful in real-life robotic experiments.

Planning techniques~\cite{Mezouar:tro:2002} can be employed to compute trajectories that achieve the desired visual task while respecting constraints. 
Alternatively, VS can be formulated as an optimization process, allowing to easily include constraints.
In~\cite{Agravante:ral:2017}, VS is written as a quadratic program (QP) so that it can account for the constrained whole-body motion of humanoid robots.
Similarly, a virtual VS written as a QP is proposed in~\cite{Paolillo:ral:2018} to achieve manipulation tasks.
%
%
Visual planning and control can be solved together using a model to predict the feature motion and the corresponding commands over a preview window~\cite{Sauvee:cdc:2006}.
Indeed, the model predictive control (MPC) technique can be applied to the VS case, by obtaining the so-called visual predictive control (VPC) framework~\cite{Allibert:tro:2010}~\cite{Allibert:book:2010}.
The main drawback of VPC is the computation time.
The flatness property~\cite{Penin:ral:2018}~\cite{Allibert:ifac:2008} can be used to reduce the problem complexity, 
but it is not applicable to all kinds of dynamics.

In this work, we propose to use a dataset of pre-processed solutions to improve VPC performance (recalled in Sect.~\ref{sec:vpc}).
To this end, an initialization and a way point is inferred on-line from the dataset.
Section~\ref{sec:soa} reports the literature on methods used to exploit stored data; the proposed approach is detailed in Sect.~\ref{sec:approach}.
Simulation and experiments, showing the effectiveness of the approach, are presented in Sect.~\ref{sec:results}. 
Section~\ref{sec:conclusion} concludes the paper and discusses future work. 


\section{BACKGROUND}\label{sec:vpc}

The VPC paradigm~\cite{Allibert:tro:2010}~\cite{Allibert:book:2010} aims at solving planning and control simultaneously. 
To this end, it computes a control sequence over a preview window by solving the optimization 
\begin{equation}
\bm{V}^\ast = \argmin_{\bm{V}\in \bm{C}} \ell(\bm{V}),
\label{eq:vpc}
\end{equation}
where the cost function is defined as 
\begin{multline}
\ell({\bm{V}}) = \sum_{j = k+1}^{k+N_p-1} ( \bm{s}_{d,j} - \bm{s}_{m,j} )^\top \bm{Q} \left( \bm{s}_{d,j} - \bm{s}_{m,j}\right) + \bm{v}_j^\top \bm{R} \bm{v}_j
\\
+ ( \bm{s}_{d,k+N_p} - \bm{s}_{m,k+N_p} )^\top \bm{Q} \left( \bm{s}_{d,k+N_p} - \bm{s}_{m,k+N_p} \right),
\label{eq:cost}
\end{multline}
and the optimization variable consists in the sequence of control actions to take along the preview window
\begin{equation}
\bm{V} = \left( \bm{v}_k,\dots,\bm{v}_{k+N_c},\dots,\bm{v}_{k+N_p-1} \right)^\top.
\label{eq:control_sequence}
\end{equation}
In~\eqref{eq:cost} and~\eqref{eq:control_sequence}, $N_p$ is the number of iterations defining the size of the preview window, while $N_c$ is the control horizon defined such that from ${k+N_c+1}$ to ${k+N_p-1}$ the control is constant and equal to $\bm{v}_{k+N_c}$; $\bm{Q}$ and $\bm{R}$ are  two matrices used to weight the error and penalize the control effort, respectively.  
In the preview window, i.e. for $j \in [k+1, k+N_p ]$, the problem is subject to
\begin{equation}
\bm{s}_{d,j} = \bm{s}^\ast - \bm{\varepsilon}_j
\quad \text{and} \quad
\bm{s}_{m,j} = \bm{s}_{m,j-1} + T_s \, \hat{\bm{L}}_j \, \bm{v}_j,
\label{eq:model}
\end{equation}
with $\bm{\varepsilon}$ the difference between the measured and the first previewed feature, constant over the preview window. $T_s$ is the sampling time.
%
Constraints on the optimization variable
\begin{equation}
\bm{V}_\text{min} \leq \bm{V} \leq \bm{V}_\text{max}
\label{eq:actuation_const}
\end{equation}
account for actuation limits, the ones on the visual features
%
\begin{eqnarray}
\bm{s}_\text{min} \leq \bm{s}_{m} \leq \bm{s}_\text{max},
\label{eq:visibility_const_a}\\
\bm{s}_m \leq \underbar{$\bm{s}$} \lor \bm{s}_m \geq \overline{\bm{s}} \label{eq:visibility_const_b},
\end{eqnarray}
%
achieve visibility constraints: \eqref{eq:visibility_const_a} forces the features to stay in an area, e.g., to prevent from leaving the image plane,~\eqref{eq:visibility_const_b} allows to avoid occlusions or spots on the lens.
The ensemble of \eqref{eq:actuation_const}-\eqref{eq:visibility_const_b} compose the set of non-linear constraints $\bm{C}$ in~\eqref{eq:vpc}.

Following the MPC rationale, at each iteration $k$, VPC measures the visual features $\bm{s}_k$, predicts the motion over the preview window using the model in~\eqref{eq:model}, minimizes the cost function~\eqref{eq:cost} and finally computes the commands $\bm{V}^\ast$. 
Only the first control of this sequence is applied to the real system which moves, providing a new set of features. 
Then, the loop starts again.
To achieve a satisfactory behavior, the control is usually kept constant over the preview window ($N_c=1$), while $N_p$ is tuned as a trade-off between a long (better tracking performance) and a short preview window (lower computational cost).
More constraints (e.g., on camera position) can be added.
In~\eqref{eq:model} a local model of the visual features is used, but a global model of the camera motion can also be considered. 
More details can be found in~\cite{Allibert:tro:2010}~\cite{Allibert:book:2010}.

Solving~\eqref{eq:vpc} with the constraints~\eqref{eq:actuation_const}-\eqref{eq:visibility_const_b} is a non-convex optimization problem.
As such, the solution depends on the solver initialization.
If it is far from the global optimum, the convergence can be slow, or get stuck in local minima providing unsatisfactory results.
Thus, it is important to provide the solver with a \emph{warm-start}, i.e., an initial commands sequence already close to the optimal solution.
To avoid the constraints, the warm-start can guide the motion away from the target momentarily. 
However, providing only warm-starts may not be sufficient.
In fact, a solver with short time horizon might consider the warm-start to be sub-optimal and modify it to move towards the goal and, as a consequence, get stuck at the local optima at the constraints.
One possible solution is to consider a long preview window and set the cost only at the end of the horizon, but this is computationally expensive.
A better idea would be to adjust the cost function with a proper \emph{way point} as sub-target to follow.


We propose to use a \emph{memory of motion}, i.e., a dataset of 
precomputed trajectories, to infer both a warm-start and a way point during the on-line VPC execution.
In this way, we leverage precomputation to shorten the VPC preview window while maintaining high performance.

\section{STATE-OF-THE-ART}\label{sec:soa}

Leveraging information stored in a memory to control or plan robotic motions has been the object of a lively research.
In~\cite{Stolle:icra:2006}, a library of trajectories is queried by k-nearest neighbor (k-NN) to infer the control action to take during the experiment.
A similar method~\cite{Liu:iros:2009} selects from the library a control which is then refined by differential dynamic programming.
As an alternative to plan from scratch, the framework in~\cite{berenson:icra:2012} starts the planner from a trajectory learned from experiences.
In~\cite{forte:ras:2012} Gaussian process regression (GPR) is used to adapt the motion, stored as dynamic motion primitives, to the actual situation perceived by the robot.
The line of works~\cite{Saveriano:icra:2014,Huber:ral:2019} considers a robot motion database built from human demonstrations.
This gives the controller a guess of the motion to make, possibly modified by the presence of obstacles.
Demonstrations and optimization techniques are used in~\cite{shen2018optimized} to handle constraints in a visual planner.

To improve the convergence of planning or control frameworks written as optimization problems, the memory can be used to provide the solvers with a warm-start.
In~\cite{Mansard:icra:2018}, a memory is iteratively built, expanding a probabilistic road map (PRM) using a local planner. 
A neural network (NN) is trained, in parallel, with the current trajectories stored in the PRM and used to give the local planner a warm-start to better connect the map.
The final NN is then used to infer the warm-start for the on-line controller.
In the context of a trajectory optimizer, the initialization is computed by applying k-NN and locally weighted regression to a set of pre-optimized trajectories~\cite{jetchev:icml:2009}.
In~\cite{Merkt:iros:2018} a k-NN infers from a memory of motion the warm-starts for a planner. 
The same kind of problem is addressed in~\cite{SantosoLembono:ral:2020} with different techniques, i.e. k-NN, GPR and Bayesian Gaussian mixture regression, that allows to also cope with multi-modal solutions. 

%
Other approaches consider the possibility to reshape the cost function to guide the solver towards an optimal solution. 
For example, the interior point method~\cite{polik2010interior} solves an inequality constrained problem by introducing the logarithmic barrier function to the cost.
In this way, the search for the solution starts from the inner region of the feasible space and then moves to the boundary region.  
In humanoid motion planning~\cite{Mordatch:ACMTG:2012}, heuristic sub-goals are introduced in the early stage of the optimization based on the zero-moment point stability criterion.
In~\cite{todorov2011convex}, to avoid discontinuity, the contact dynamics are smoothened such that virtual contact forces can exist at a distance.
%
In reinforcement learning, it is common to modify the sparse reward function, that is difficult to achieve, by providing intermediate rewards as way points~\cite{peng2018deepmimic}. 

To build our framework and successfully achieve VPC tasks, we took inspiration from the different approaches existing in the literature.
In particular, we decided to exploit the information contained in a memory of motion to infer: (i) warm-start to well initialize our optimization solver; and (ii) way point to be used in the cost in lieu of the final target.

\section{THE PROPOSED APPROACH}\label{sec:approach}

As recalled in Sect.~\ref{sec:vpc}, VPC computes a control sequence $\bm{V}^\ast$ by solving a minimization problem. 
To efficiently find an optimal solution, the process has to converge fast and avoid local minima.
Thus, it is important to initialize the solver with a warm-start $\bm{V}_\text{ini}$, and reshape the cost function using a way point $\bar{\bm{s}}$ in place of the target $\bm{s}^\ast$.
%
%
This section explains how to infer the warm-start and way point from a memory. 

The memory of motion is a dataset ${\cal D}= \{ (\bm{x}_i$,$\bm{y}_i) \}$ of $D$ samples.
Each feature $\bm{x}_i$ describes a particular visual configuration and is composed of a set $\bm{s}_i$ of visual features, the area $a_i$ and the orientation $\alpha_i$ of the visual pattern\footnote{For example, if point features are used, the visual pattern is the polygon having the visual features as vertexes.}
\begin{equation}
\bm{x}_i = \left( \bm{s}_i^\top, a_i, \alpha_i \right)^\top \in \mathbb{R}^{n \times 1},
\label{eq:reg_feature}
\end{equation}
where $n = n_f + 2$, $n_f$ is the dimension of the visual feedback $\bm{s}$.
We consider $a_i$ and $\alpha_i$ along with $\bm{s}_i$ in~\eqref{eq:reg_feature}  to make the samples distinguishable, not only in terms of the visual appearance but also w.r.t. the corresponding camera poses.
The output variable $\bm{y}_i$ contains the proper control action to take and the way point to follow in function of $\bm{x}_i$.
Since the control is constant in the preview window ($N_c = 1$, see Sect.~\ref{sec:vpc}), it is enough to store the single command 
\begin{equation}
\bm{y}_i = \left( \bm{v}_i^\top, {\bar{\bm{s}}_i}^\top \right)^\top \in \mathbb{R}^{p \times 1},
\label{eq:reg_output}
\end{equation}
where $p = q + n_f$, with $q$ the actuated degrees of freedom of the camera.
All the samples are collected in the matrices
\begin{equation}
\bm{X} = 
\left(
\begin{array}{c}
\bm{x}_1^\top\\
\vdots\\
\bm{x}_{D}^\top
\end{array}
\right) \in \mathbb{R}^{D \times n}
~, ~
\bm{Y} =
\left(
\begin{array}{c}
\bm{y}_1^\top\\
\vdots\\
\bm{y}_{D}^\top
\end{array}
\right)  \in \mathbb{R}^{D \times p}.
\label{eq:reg_matrices}
\end{equation}
The whole process computing warm-start and way point consists in off-line building and on-line querying the memory.

\begin{figure}[t!]
\centering
\vspace{2mm}
\begin{algorithmic}
\small
\STATE {\bf Input}: $N_t, \bm{s}^\ast$ {\bf Output}: $\bm{X}$, $\bm{Y}$ 
\vspace{-2mm}
\\\hrulefill
\STATE $\bm{X}=\{\,\}, \bm{Y} = \{\,\}, n_t = 0$
\WHILE{$n_t < N_t$}
\STATE $\bm{v}_0 = \bm{0}$
\STATE $\bm{s}_0 \leftarrow \textsc{Generate Initial Visual Features}$
\STATE $L=0$, \texttt{success}~$=$~\texttt{False}
\WHILE{$\neg (\bm{s}_L \rightarrow \bm{s}^\ast) \land (\texttt{success}~\text{is}~\texttt{True}) $ }
\STATE $\bm{v}_L, \texttt{success} \leftarrow \textsc{Find Solution}(\bm{s}_L,\bm{s}^\ast,\bm{v}_{L-1})$
\STATE $\bm{s}_{L+1} \leftarrow \textsc{Update Model}(\bm{s}_L,\bm{v}_{L})$
\STATE $L = L+1$
\ENDWHILE
\IF{\texttt{success} is \texttt{True}}
\STATE $n_t = n_t + 1$
\FOR{$j = 0,1, \dots, L$}
\STATE $a_j, \alpha_j \leftarrow \textsc{Compute Area and Angle}(\bm{s}_j)$
\STATE $\bar{\bm{s}}_j \leftarrow \textsc{Compute Way Point}(\bm{s}_j)$
\STATE $\bm{x}_j = ( \bm{s}_j^\top, a_j, \alpha_j )^\top, \bm{y}_j = ( \bm{v}_j^\top, {\bar{\bm{s}}_j}^\top )^\top$
\STATE \textsc{Store} $\bm{x}_j$ and $\bm{y}_j$ in $\bm{X}$ and $\bm{Y}$
\ENDFOR
\ENDIF
\ENDWHILE
\end{algorithmic}
\caption{Algorithm generating the trajectories for the memory of motion.}
\label{fig:algo_memory_building}
\end{figure}

\subsection{Building the memory of motion}\label{sec:building_memmo}

The memory of motion is built by running VPC off-line for different sets of initial visual features.
The aim is to compute successful trajectories able to achieve the visual task.
To this end, the same solver of the on-line executions is used to build the memory.
However, since the aim is to build `high-quality' samples and there is no strict constraint on the execution time (the memory is built off-line), the solver is set up with low thresholds on the solution optimality, a high number of max iterations allowed, and a large VPC preview window.

The process building the memory of motion is presented in the algorithm of Fig.~\ref{fig:algo_memory_building}.
For $N_t$ random initial conditions $\bm{s}_0$, if the VPC solver succeeds to find a feasible solution (no constraint is violated) and the task is achieved ($\bm{s}$ converge to $\bm{s}^\ast$ in the given time), then all the visual features from $\bm{s}_0$ to $\bm{s}_L$ are saved ($L$ is the length of the trajectory). 
Thus, $\forall \, \bm{s}_j, j= 0,\dots,L$, the following actions are executed:
\begin{itemize}
\item the area $a_j$ and angle $\alpha_j$ of the corresponding visual pattern $\bm{s}_j$ are computed;
\item the way point is computed as the visual features at $n_\text{s}$ samples ahead ($\bar{\bm{s}}_j = \bm{s}_{j+n_\text{s}}$); if $j+n_\text{s}>L$, $\bar{\bm{s}}_j=\bm{s}^\ast$;
\item the corresponding solution $\bm{v}_j$ is selected.
\end{itemize}
With this information, the vectors $\bm{x}_j$ and $\bm{y}_j$ are obtained and finally stored in $\bm{X}$ and $\bm{Y}$.
The initial value of the visual features $\bm{s}_0$ is generated randomly at the start of the memory building, while at the later stage it is biased toward the distributions corresponding to the set of unsuccessful initial conditions (estimated by Gaussian Mixture Model), so that the solver attempts to solve the difficult cases when the database has contained a sufficient number of samples.
%
The algorithm uses the function `\textsc{Find Solution}' which tries to find an optimal solution, employing the strategies detailed in the algorithm of Fig.~\ref{fig:algo_find_solution}.
It implements an iterative mechanism by which the memory building process benefits from the current status of the memory itself.
Indeed, if there are enough trajectories in the memory, and the features are close to the constraints (in which case the function `\textsc{Is\_Close}' returns \text{True}), the solver is provided with a warm-start and way point inferred by a k-NN algorithm (details in Sect.~\ref{sec:query_memmo}).
Otherwise, the algorithm tries to solve the VPC using the previous solution $\bm{v}_{L-1}$ as warm-start.
If the solver does not manage to find a successful solution, two recovery strategies are executed: the solver is warm-started with one of: (i) 12 pre-defined; or (ii) 10 random camera velocity directions.
%
%
%
In the presented algorithms, `$\land$', and `$\lnot$' denote the AND and NOT logic operator, respectively.  
Once the memory is built, it is ready to be queried on-line.

\begin{figure}[t!]
\centering
\vspace{2mm}
\begin{algorithmic}
\small
\STATE {\bf Input}: $\bm{s}_L$, $\bm{s}^\ast$, $\bm{v}_{L-1}$, $n_t$ $\bm{X}$, $\bm{Y}$ {\bf Output}: $\bm{v}_L$, \texttt{success} 
\vspace{-2mm}
\\\hrulefill
\IF{\textsc{Is\_Close}($\bm{s}_L$) is \texttt{True} $\land$ $n_t~> n_\text{th}$}
\STATE $a_L, \alpha_L \leftarrow \textsc{Compute Area and Angle}(\bm{s}_L)$
\STATE $\hat{\bm{x}} = (\bm{s}_L^\top, a_L, \alpha_L )^\top$
\STATE $\hat{\bm{v}}, \bar{\bm{s}} \leftarrow \textsc{k-NN}(\bm{X},\bm{Y},\hat{\bm{x}})$
\STATE $\bm{v}_L, \texttt{success} \leftarrow \textsc{Solve VPC}(\bm{s}_L,\bar{\bm{s}},\hat{\bm{v}})$
\ELSE
\STATE $\bm{v}_L, \texttt{success} \leftarrow \textsc{Solve VPC}(\bm{s}_L,\bm{s}^\ast,\bm{v}_{L-1})$
\ENDIF
\STATE $k = 0$
\WHILE{\texttt{success} is \texttt{False} $\land$ $k < 12$}
\STATE $\bm{v}_\text{dir} \leftarrow \textsc{Select Direction}(k)$ 
\STATE $\bm{v}_L, \texttt{success} \leftarrow \textsc{Solve VPC}(\bm{s}_L,\bm{s}^\ast,\bm{v}_\text{dir})$
\STATE $k = k+1$
\ENDWHILE
\STATE $k = 0$
\WHILE{\texttt{success} is \texttt{False} $\land$ $k < 10$}
\STATE $\bm{v}_\text{rand} \leftarrow \textsc{Select Random}(k)$ 
\STATE $\bm{v}_L, \texttt{success} \leftarrow \textsc{Solve VPC}(\bm{s}_L,\bm{s}^\ast,
\bm{v}_\text{rand})$
\STATE $k = k+1$
\ENDWHILE
\end{algorithmic}
\caption{Algorithm trying to find a feasible VPC solution.}
\label{fig:algo_find_solution}
\end{figure}

\subsection{Querying the memory of motion}\label{sec:query_memmo}

The aim of querying the memory of motion is to infer from the dataset proper initial guess and way point for the on-line VPC solver, given the current visual features configuration.
This means that we need to learn the map $\bm{f} : \bm{x} \to \bm{y}$ from ${\cal D}$ so that an estimate $\hat{\bm{y}}$ can be computed for a novel feature $\hat{\bm{x}}$.
The map $\bm{f}$ is learned using standard regression techniques, i.e. k-NN and GPR, as also proposed in~\cite{SantosoLembono:ral:2020}. 
In what follows, we describe the adaptation required for the VPC application.
%

The k-NN algorithm is a simple non-parametric method selecting the $K$ closest samples in the dataset ${\cal D}$, given a new feature $\hat{\bm{x}}$. The distance between samples is computed as Euclidean norm.
The corresponding $K$ closest outputs are thus averaged to provide the estimated output
\begin{equation}
\hat{\bm{y}} = \frac{1}{K} \sum_{i=1}^K \bm{y}_i.
\label{eq:knn}
\end{equation}
In the case of GPR~\cite{Rasmussen:book:2006}, the inference is computed by 
\begin{equation}
\hat{\bm{y}}= \bm{m} + \bm{K}(\hat{\bm{x}}, \bm{X}) \bm{\varLambda}(\bm{X},\bm{Y}),
\label{eq:gpr}
\end{equation}
where $\bm{\varLambda} = \bm{K}(\hat{\bm{x}}, \bm{X})  \left( \bm{K}(\bm{X},\bm{X} ) + \varphi_s \bm{I} \right)^{-1} (\bm{Y} - \bm{m})$  can be computed off-line, so that only a vector sum and a matrix multiplication, fast to compute, are left for the on-line estimation; $\bm{I}$ is the identity matrix\footnote{Hereafter, $\bm{I}$, $\bm{1}$ and $\bm{0}$ refer to the identity, all-ones and null matrix. When not explicitly marked, the dimensions are inferred from the context.}.
In~\eqref{eq:gpr}, $\bm{K}$ is the covariance matrix which is built from the kernel function.
A popular choice, also used in this work, is the radial basis function 
$\bm{K}_{ij} =  \varphi_o^2 \exp \left( -\tfrac{1}{2}(\bm{x}_i -\bm{x}_j)^\top  \bm{\varPhi} (\bm{x}_i - \bm{x}_j) \right)$. %
%
%
The hyperparameters $\bm{\varPhi}$, $\varphi_0$ and $\varphi_s$ are computed by minimizing the marginal log-likelihood. 
Finally, $\bm{m}$ is the mean function acting as an offset in the estimation process. 
We consider the constant vector $\bm{m} = ( \bm{0}_{1\times q}, \, \bm{s}^{\ast \top} )^\top$
%
that suggests to compute zero velocity as warm-start and the final target as way point when $\hat{\bm{x}}$ is in an area not sampled by the memory.
GPR is known to be effective with small data-set and is fast to compute.
These characteristics fit very well our task, since the memory is built with trajectories lying on the image plane (which is a limited area) and has to be queried fast to be compatible with the on-line control requirements.

Finally, recalling that the control is constant in the preview window, the warm-start is built from the first $q$ entries of $\hat{\bm{y}}$: 
\begin{equation}
\bm{V}_\text{ini} = (\hat{y}_1, \dots, \hat{y}_q) \otimes \bm{1}_{1\times (N_p -1)},
\label{eq:warm_start}
\end{equation}
where `$\otimes$' is the Kronecker product.
The way point, instead, is obtained from the remaining $n_f$ elements of $\hat{\bm{y}}$:
\begin{equation}
\bar{\bm{s}} = (\hat{y}_{q+1}, \dots, \hat{y}_{q+n_f})^\top.
\label{eq:way_point}
\end{equation}

Note that in the absence of constraints, 
%
the solution found at the previous iteration is already a good warm-start for the solver and there is no need to reshape the cost with a way point.
%
Thus, the memory-based strategy is activated only when the visual features are ``close'' to one of the visibility constraints, i.e., when the distance between any feature and the border of the constraints is lower than a given threshold.

\begin{figure}
\vspace{1mm}
\centering
\includegraphics[width=0.3\textwidth]{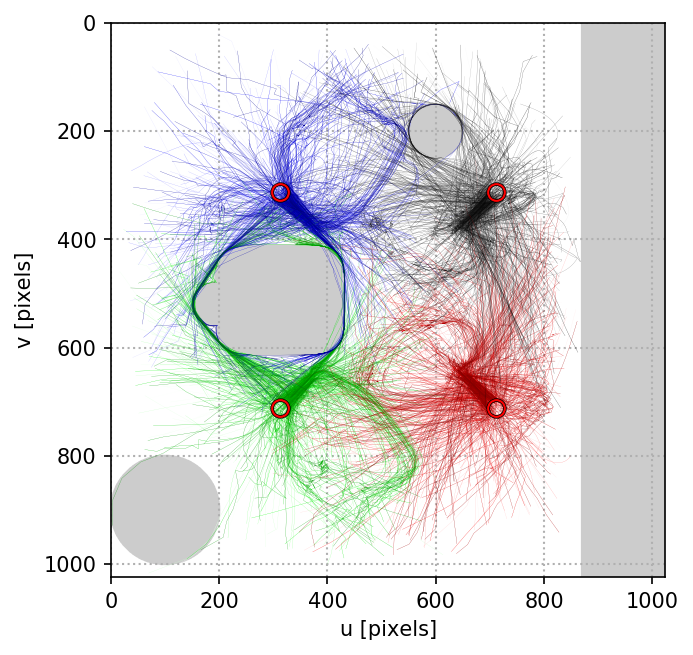}
\caption{Simulations: visual features trajectories stored in the memory. For visualization purposes, the samples are plotted with different shades; each color corresponds to the motion of one single feature.}
\label{fig:memmo}
\end{figure}

\section{RESULTS}\label{sec:results}

In this section we present the results carried out with the proposed framework to efficiently achieve VPC tasks.

As visual features $\bm{s}$, we considered four points ($n_f = 8$). The visual task consisted in making them match with four corresponding desired points $\bm{s}^\ast$.
The image Jacobian in~\eqref{eq:model} has been approximated using the points depth at the target, known in advance.
The approach has been implemented in Python. As optimization solver, we used the SLSQP method available in the open source library SciPy~\cite{scipy}.
Actuation and visibility constraints were implemented as bounds and non-linear inequality constraints.
The OR logic operation in~\eqref{eq:visibility_const_b}, to be implemented, was converted into AND with a $p$-norm formulation~\cite{Hyun:ral:2017}.
We choose $K=1$ for our k-NN, that is thus mainly used to select samples as they are in the memory; we considered the GPy library~\cite{gpy} as GPR implementation. 
As explained in Sect.~\ref{sec:vpc}, VPC was set-up with $N_c=1$ and $T = 1/30$~s (since $30$~Hz is the camera nominal framerate).

\subsection{Simulations}
\begin{figure*}[t!]
\centering
\subfloat[Prev.-it. strategy: visual feature path]{\includegraphics[width=0.28\textwidth]{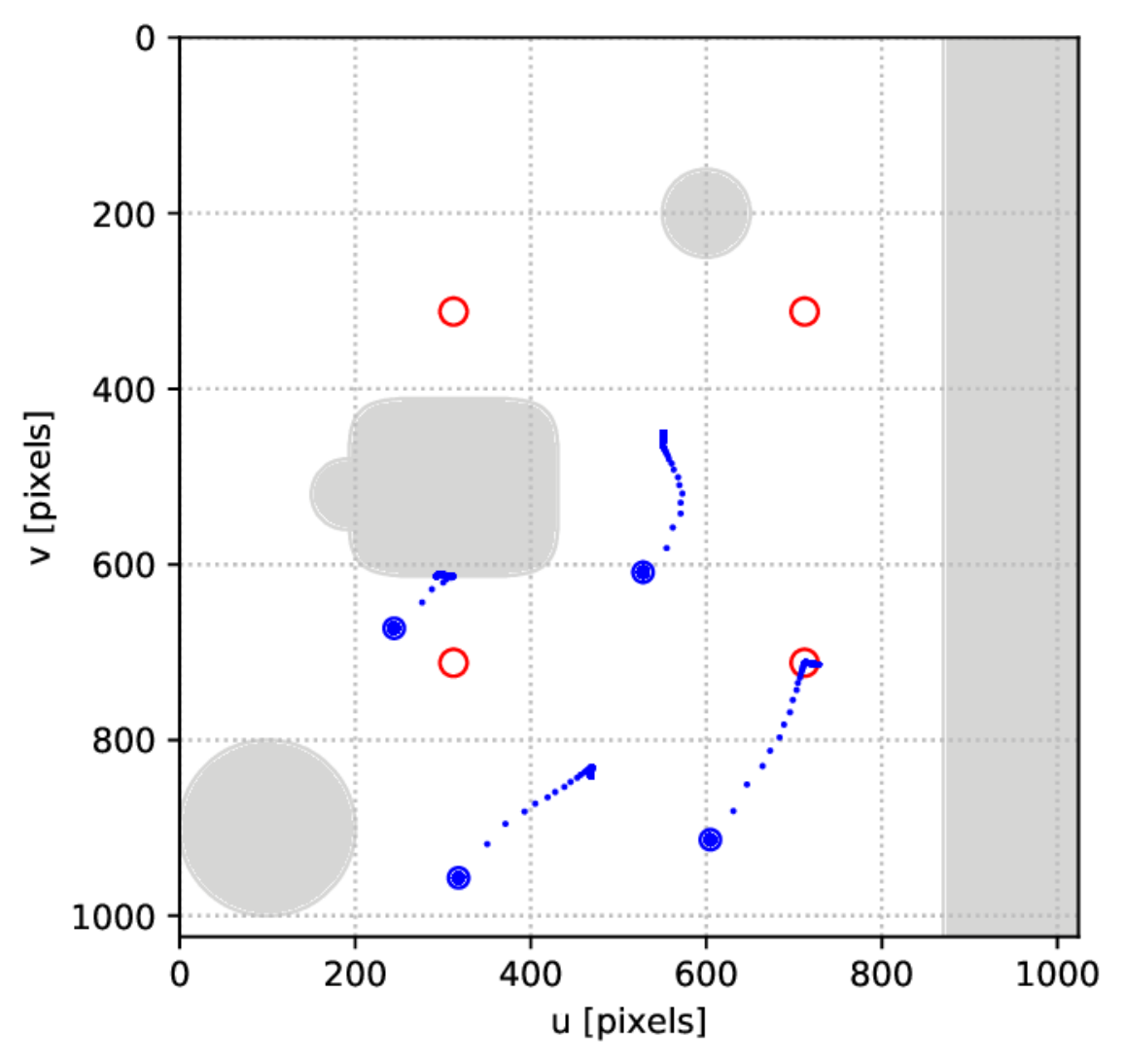}\label{fig:image_plane_previt}}
\hfill
\subfloat[k-NN-based strategy: visual feature path]{\includegraphics[width=0.28\textwidth]{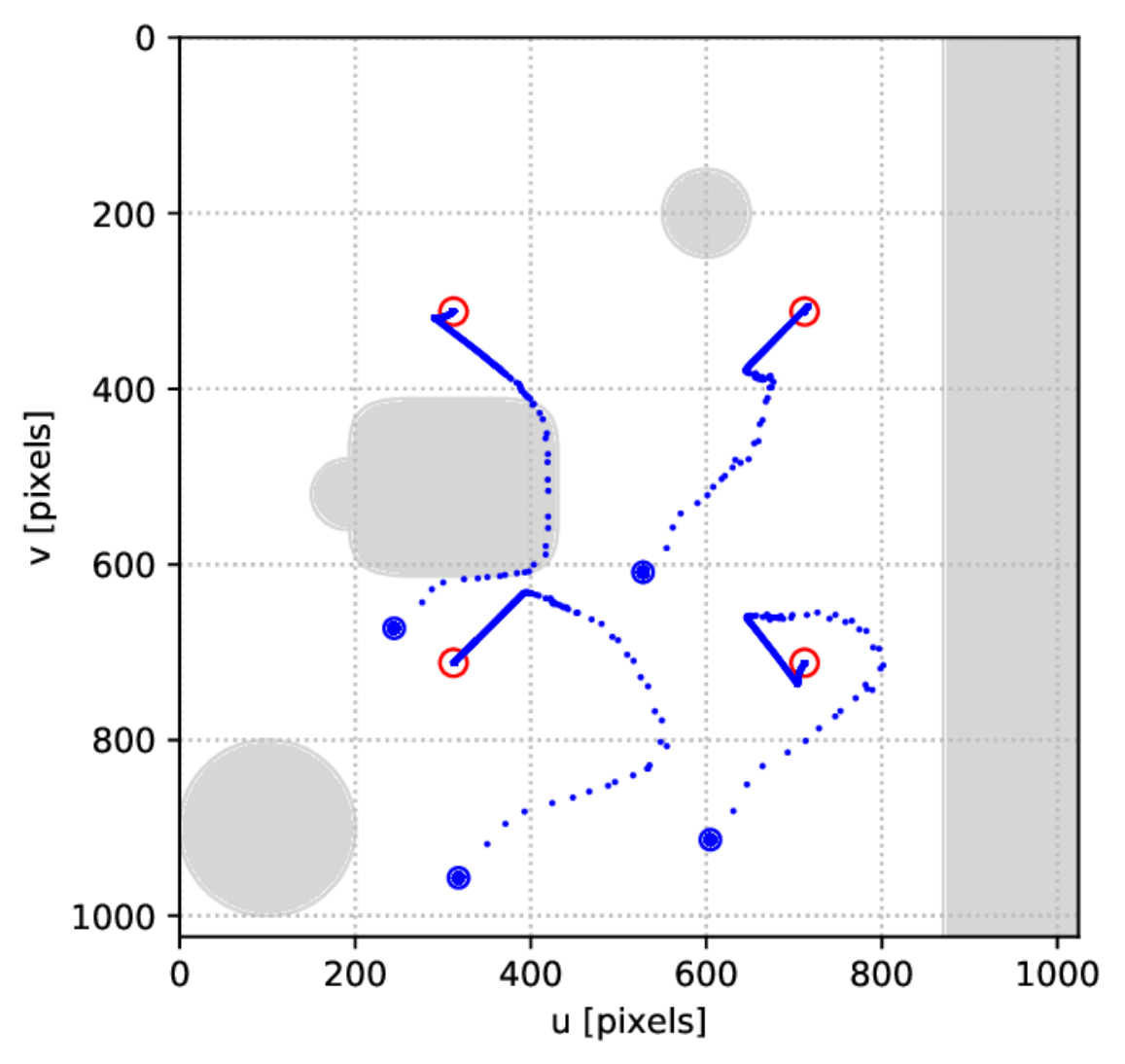}\label{fig:image_plane_kNN}}
\hfill
\subfloat[GPR-based strategy: visual feature path]{\includegraphics[width=0.28\textwidth]{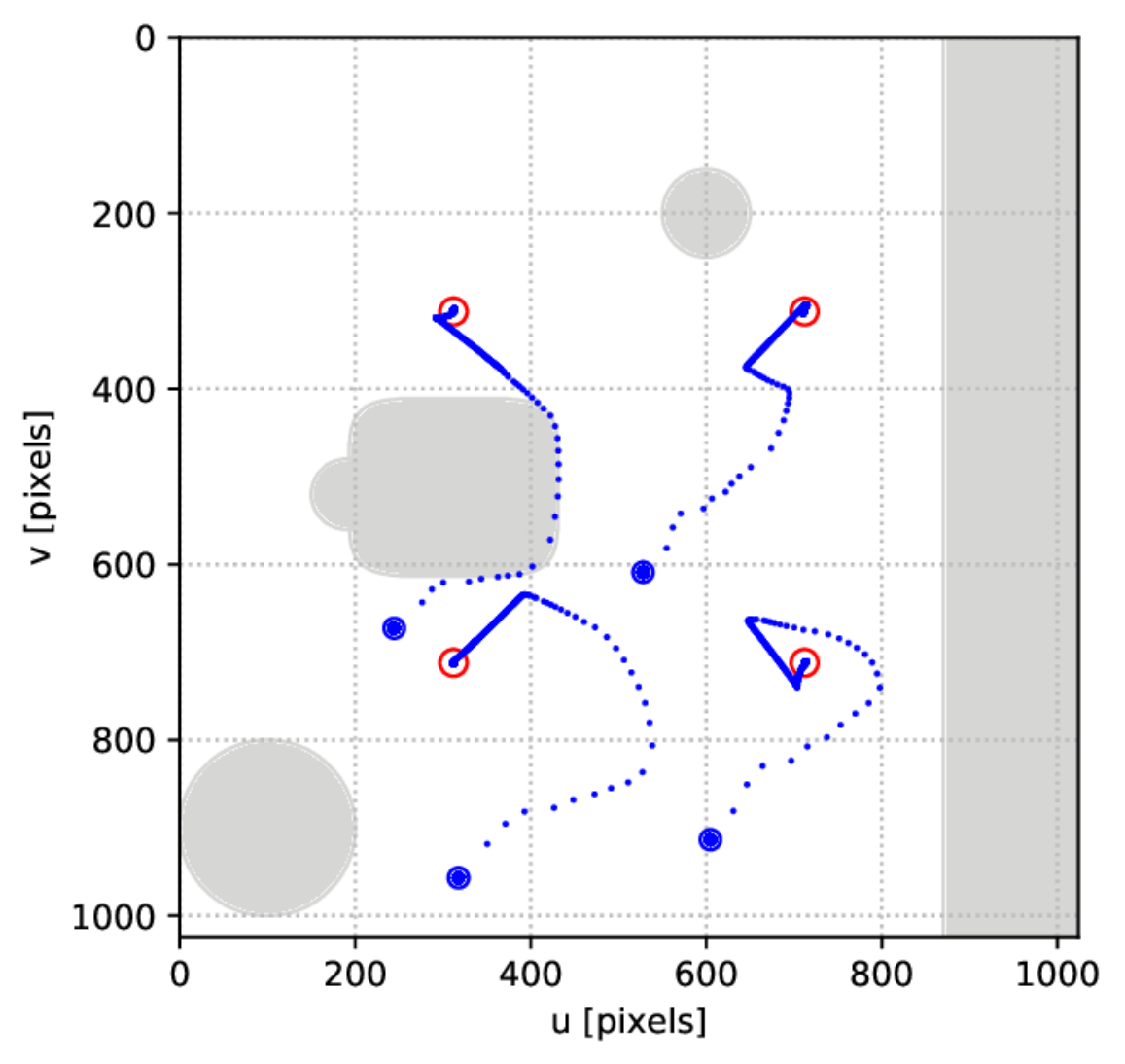}\label{fig:image_plane_GPR}}
\\
\subfloat[Prev.-it. strategy: velocity command]{\includegraphics[width=0.28\textwidth]{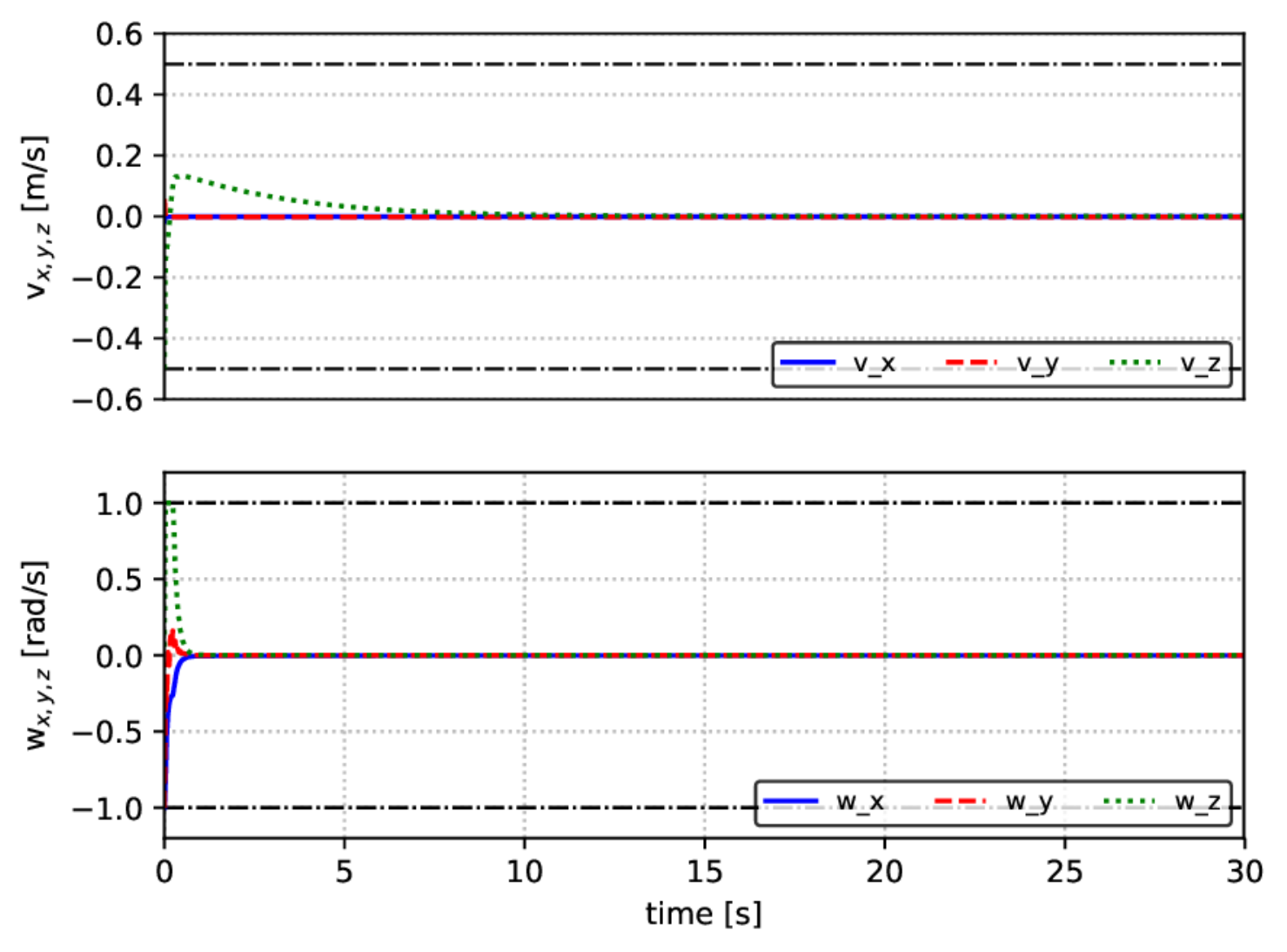}\label{fig:vel_cmd_previt}}
\hfill
\subfloat[k-NN-based strategy: velocity command]{\includegraphics[width=0.28\textwidth]{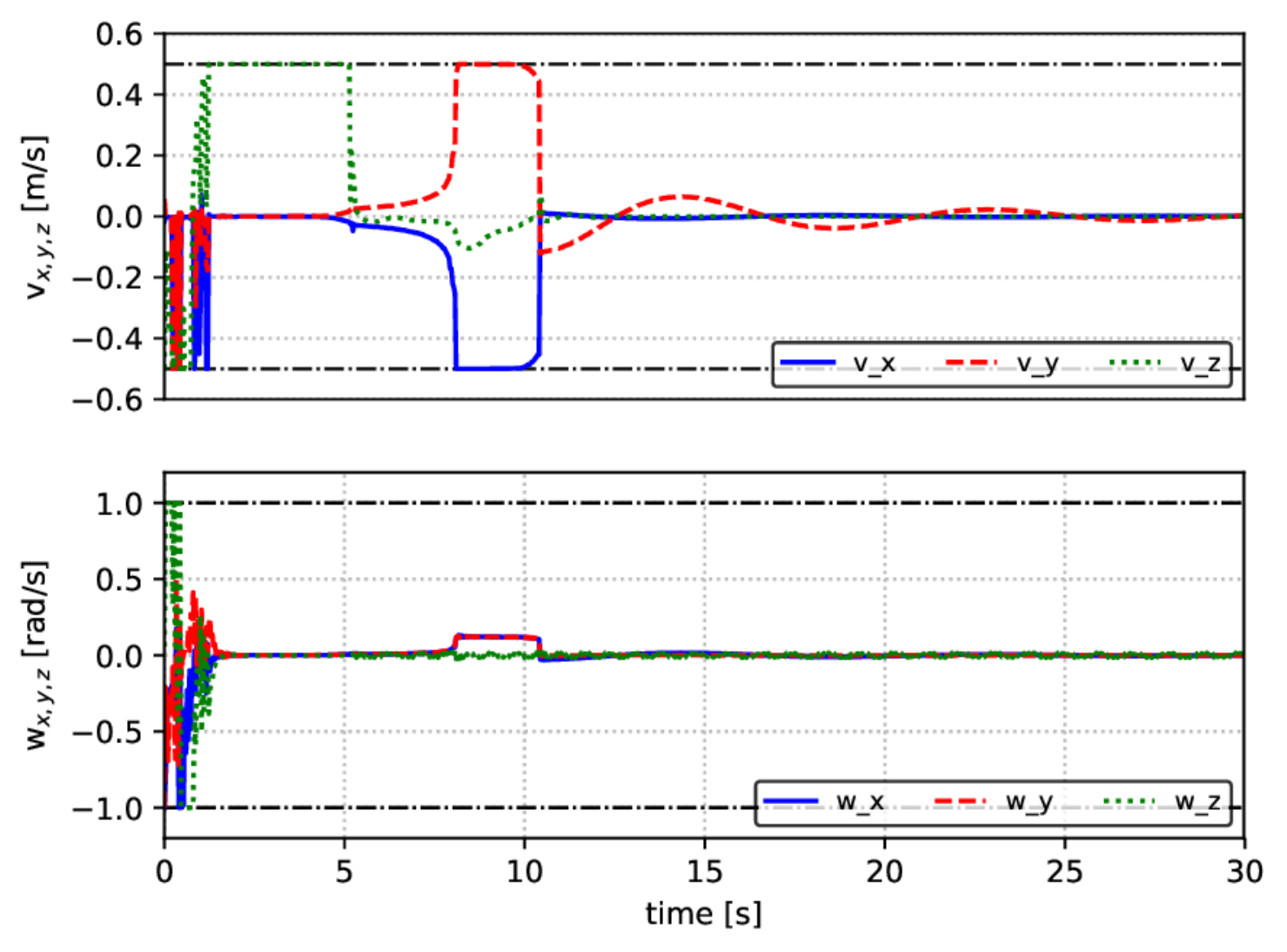}\label{fig:vel_cmd_kNN}}
\hfill
\subfloat[GPR-based strategy: velocity command]{\includegraphics[width=0.28\textwidth]{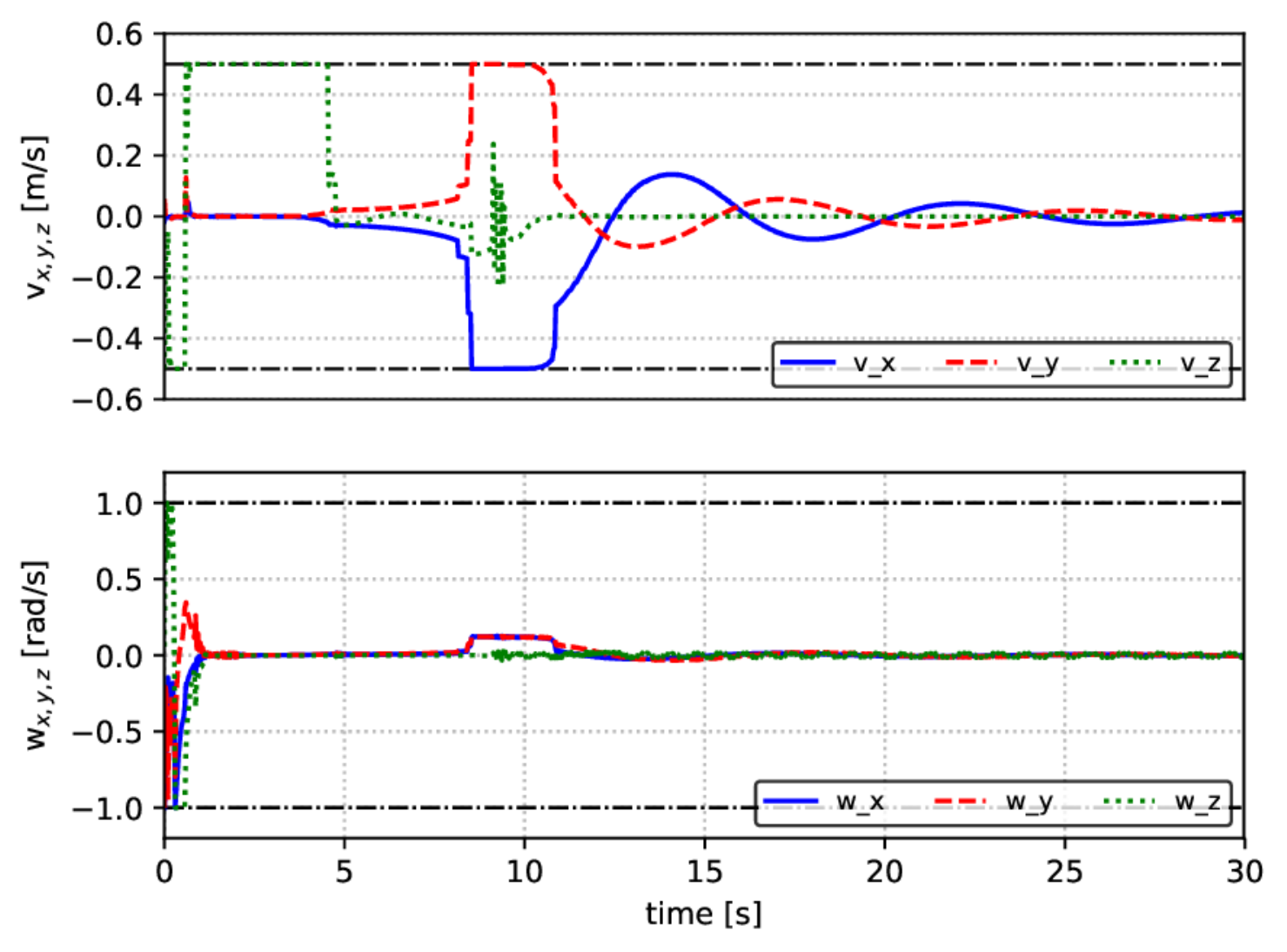}\label{fig:vel_cmd_GPR}}
\caption{Simulations: comparison between prev.-it., k-NN and GPR-based strategies in terms of features motion (top) and velocity command (bottom).}
\label{fig:simulation_one_shot}
\end{figure*}
For the simulations, we considered a hand-held camera free to move in the Cartesian space ($q=6$), with an image resolution of $1024\times 1024$~pixels.
%
As visibility constraints, we considered four convex and concave areas on the image ($u$-$v$ plane) simulating occlusions and spots on the lens.
%
As actuation constraints, we limited the linear and angular velocity components of the camera to $\pm 0.5$~m/s and $\pm 1$~rad/s.
We set $\bm{R} = \diag(100, 100, 1,0.5, 0.5, 0.5)$  (decreasing to $\bm{0}$ towards the convergence) and $\bm{Q} = k_Q \bm{I}$ with $k_Q=10^{-3}$.
The memory of motion was generated following the procedure of Sect.~\ref{sec:building_memmo}. 
In particular, the solver was set-up with an optimality precision of $10^{-9}$ and $100$ maximum iterations. VPC was set with $N_p=10$.
The choice of these parameters was driven by the need to store `high-quality' samples, at the cost of a high computational time that we were willing to pay since the memory is built off-line.
We generated $N_t=900$ trajectories, for a total of $D=16010$ samples.
Fig.~\ref{fig:memmo} shows the visual features trajectories stored in the memory.
The visibility constraints are depicted as shadowed areas, while the target $\bm{s}^\ast$ are the red circles.
%
%
%
For the on-line executions, we relaxed the solver parameters with $10^{-3}$ as optimality precision and $10$ maximum iterations.
This set-up, along with a smaller $N_p$, allowed faster computations.
However, thanks to the memory-based strategies presented in Sect.~\ref{sec:approach}, performances are not invalidated, but even improved. 

\begin{table}[b]
\centering
\caption{Simulations: statistics comparing different VPC strategies.}
\begin{tabularx}{\columnwidth}{l c X c X c}
\toprule
\textbf{Strategy} & $r$ (\%) & & $\bar{t}_c$~(s) & & $\bar{\ell}$ \\  \midrule
\textbf{Previous-iteration ($N_p = 3$)}   	& 80 & &  0.085 & & 49.3 \\
\textbf{Previous-iteration ($N_p = 30$)}   & 83 & & 0.550 & & 52.9  \\
\textbf{k-NN-based} & 92 	& & 0.074 & & 19.6 \\
\textbf{GPR-based}  & 93	& & 0.080 & & 16.4 \\ 
\bottomrule
\end{tabularx}
\label{tab:statistical}
\end{table}
The approach was first evaluated with a statistical analysis, comparing: VPC warm-started with the previous-iteration solution (for brevity denoted ``prev.-it.'') (i) using $N_p=3$ and (ii) using $N_p=30$; using warm-start and way point provided (iii) by k-NN and (iv) by GPR, both with $N_p=3$.
The memory-based strategies were activated at $20$~pixels from the occlusions and we set $n_s = 5$. 
For GPR, data were sub-sampled by a factor $20$.
The comparison is performed w.r.t. the success rate $r$, the average of the solver convergence time $\bar{t}_c$ and the average of the cost $\bar{\ell}$ divided by $N_p$ for all (successful and unsuccessful) trajectories.
Each execution is considered successful if no constraint is violated (with a tolerance of $15$~pixels) and the visual task is achieved ($\bm{s}$ converges to $\bm{s}^\ast$ in the given time of $15$~s).
Each strategy was tested using the same $100$ random initial configuration. 
The results, run on a laptop with an i7-$1.80$~GHz 4-cores and $8$~GiB RAM, are reported in Table~\ref{tab:statistical}.
The prev-it strategy with $N_p=3$ allowed to obtain $80$\% of success rate (note that among the test samples, many had an easy task execution).
In order to improve $r$, for the considered scenario, we had to increase the preview window to $N_p = 30$, but this also increased the computation time.
The proposed memory-based strategies allowed us to keep the preview window short, so that both $\bar{t}_c$ and $\bar{\ell}$ have low values, and increase $r$ at the same time.
This is due to the effect of warm-start and way point which help the execution of the task.

The main reason of the prev.-it. strategy failures is that the solution gets stuck at the visual occlusions.
The memory-based strategies reduces the occurrence of these situations. 
As an example, in Fig.~\ref{fig:simulation_one_shot} we present the plots related to a single task execution, where the big blue dot is the initial value of the features, the smaller blue dots are the VPC solutions at each iteration, whereas the red circles are the target.
The prev.-it. strategy stops at an occlusion border (see Fig.~\ref{fig:image_plane_previt}), as effect of conflicting gradients that produce zero velocity commands (Fig.~\ref{fig:vel_cmd_previt}).
Instead, the memory-based approaches manage to overcome the occlusion, as shown in Figs~\ref{fig:image_plane_kNN} and~\ref{fig:image_plane_GPR}.
In particular, the GPR solution, thanks to its interpolation capabilities, produces a smoother behavior w.r.t. our k-NN implementation (cf. Figs~\ref{fig:image_plane_kNN}-\ref{fig:vel_cmd_kNN} with~\ref{fig:image_plane_GPR}-\ref{fig:vel_cmd_GPR} ).
%

\subsection{Robot experiments}
\begin{figure}[t]
\begin{center}
\subfloat[]{\includegraphics[width=0.32\columnwidth]{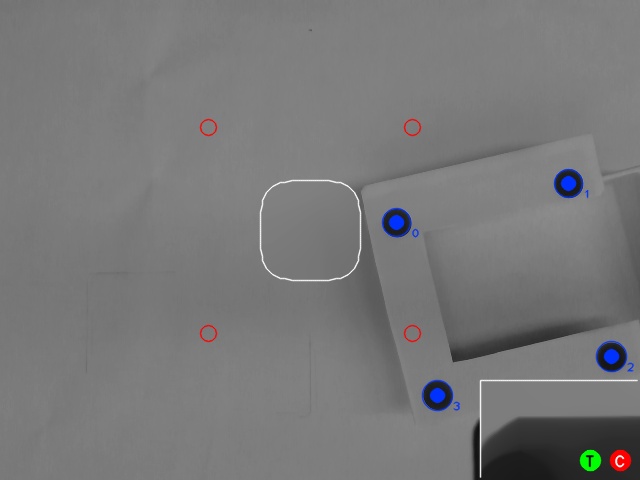}\label{fig:exp_fail_a}}
\hfill
\subfloat[]{\includegraphics[width=0.32\columnwidth]{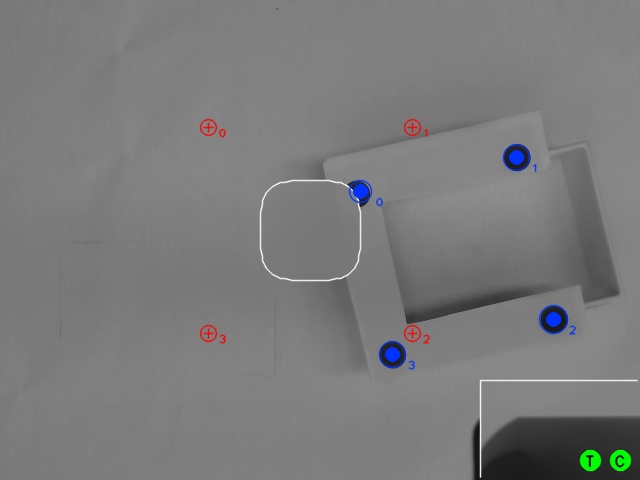}\label{fig:exp_fail_b}}
\hfill
\subfloat[]{\includegraphics[width=0.32\columnwidth]{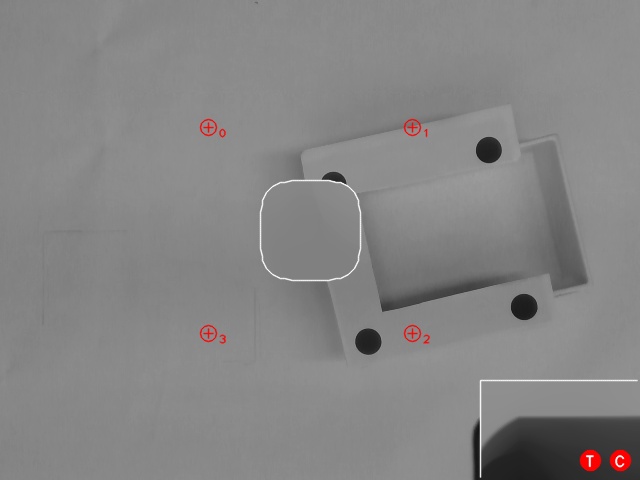}\label{fig:exp_fail_c}}
\end{center}
\caption{Robot experiment: the prev.-it. strategy fails to avoid the occlusion.}
\label{fig:exp_fail}
\end{figure}
For the experiments, we used the 7 degrees-of-freedom robot arm Panda by Franka Emika, with an Intel RealSense RGB-D sensor mounted at the end-effector.
The sensor, used as monocular camera, outputs images with a resolution of $640\times480$~pixels at a nominal framerate of $30$~Hz.
The image processing, used to detect the point features, was implemented using the open source library OpenCV~\cite{opencv}. 
A calibration procedure computed the intrinsic camera parameters and the camera--end-effector displacement.
The camera velocity commands, computed by VPC, were transformed in the robot frame and sent to the robot Cartesian controller. 
As task, the robot had to place an object inside a box where we placed four known markers.
Without knowing the box pose, VPC was used to drive the robot over the box and, after convergence, release the object.
On the image we considered two constraints to take into account the occlusion of the object grasped by the robot, and emulate a spot in the center of the lens as a blurred area.
VPC was set with $k_Q=10^{-6}$, $\bm{R} =\diag(50, 50, 0.5, 0.25,  0.25, 0.25)$ (decreasing it to $0.1 \bm{R}$ approaching the convergence), while the commands bounds were set to $\pm0.01$~m/s and $\pm0.03$~rad/s.


The memory ($450$ trajectories for a total of $64339$ samples) was built with $N_p=5$, solver optimality tolerance of $10^{-9}$ and $50$ maximum iterations.
The iterative building and the adaptive sampling were not used.
To be conservative, the spot considered in the memory was bigger than the one in the experiments.
Given the simulation results, we decided to use the GPR-based strategy.
Data were subsampled by a factor $60$, with $n_\text{s} = 30$.
The trigger signal to query the memory was activated at $30$~pixels from the occlusions.

\begin{figure}[t!]
\begin{center}
\subfloat[]{\includegraphics[width=0.32\columnwidth]{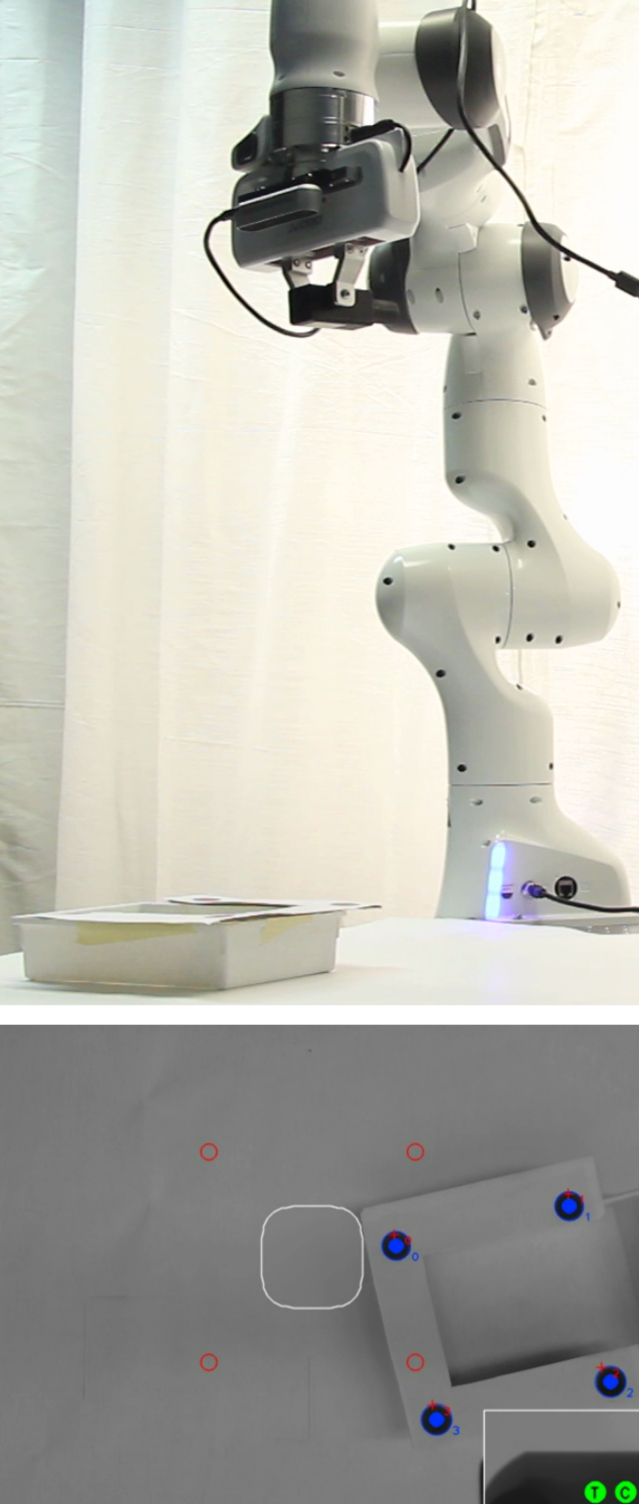}\label{fig:exp_a}}
\hfill
\subfloat[]{\includegraphics[width=0.32\columnwidth]{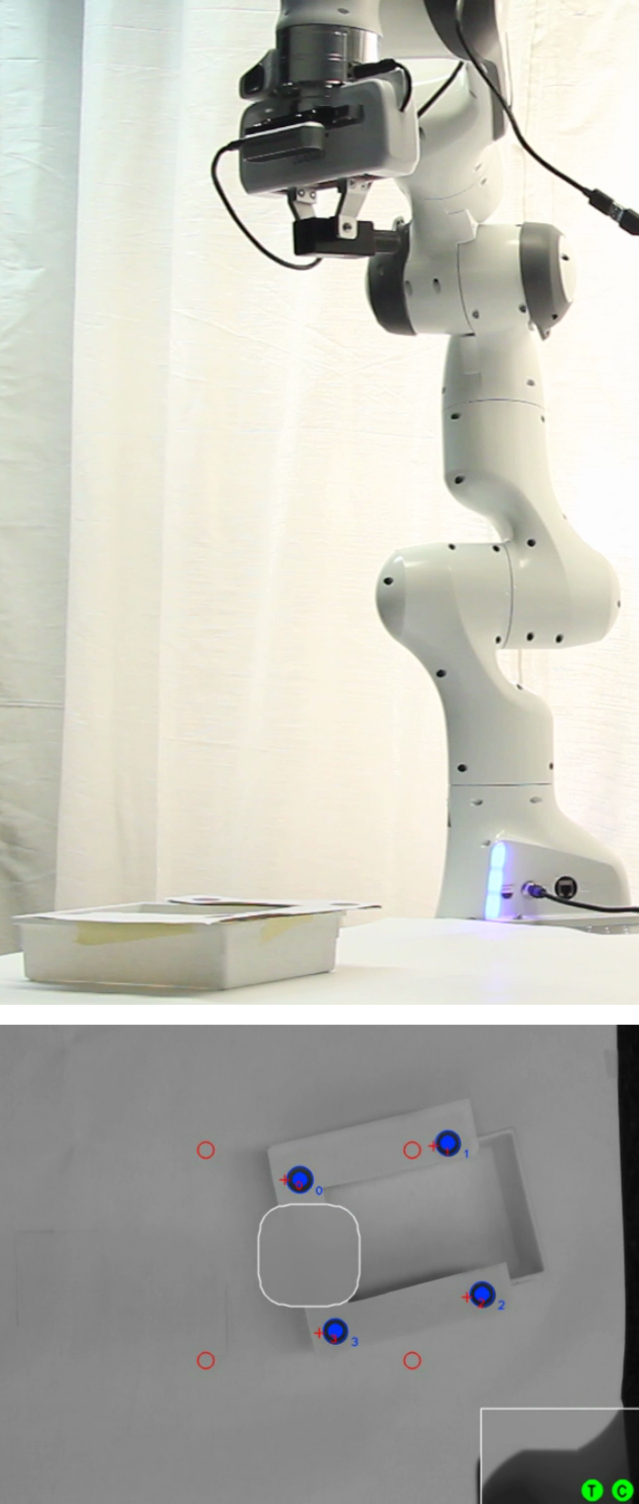}\label{fig:exp_b}}
\hfill
\subfloat[]{\includegraphics[width=0.32\columnwidth]{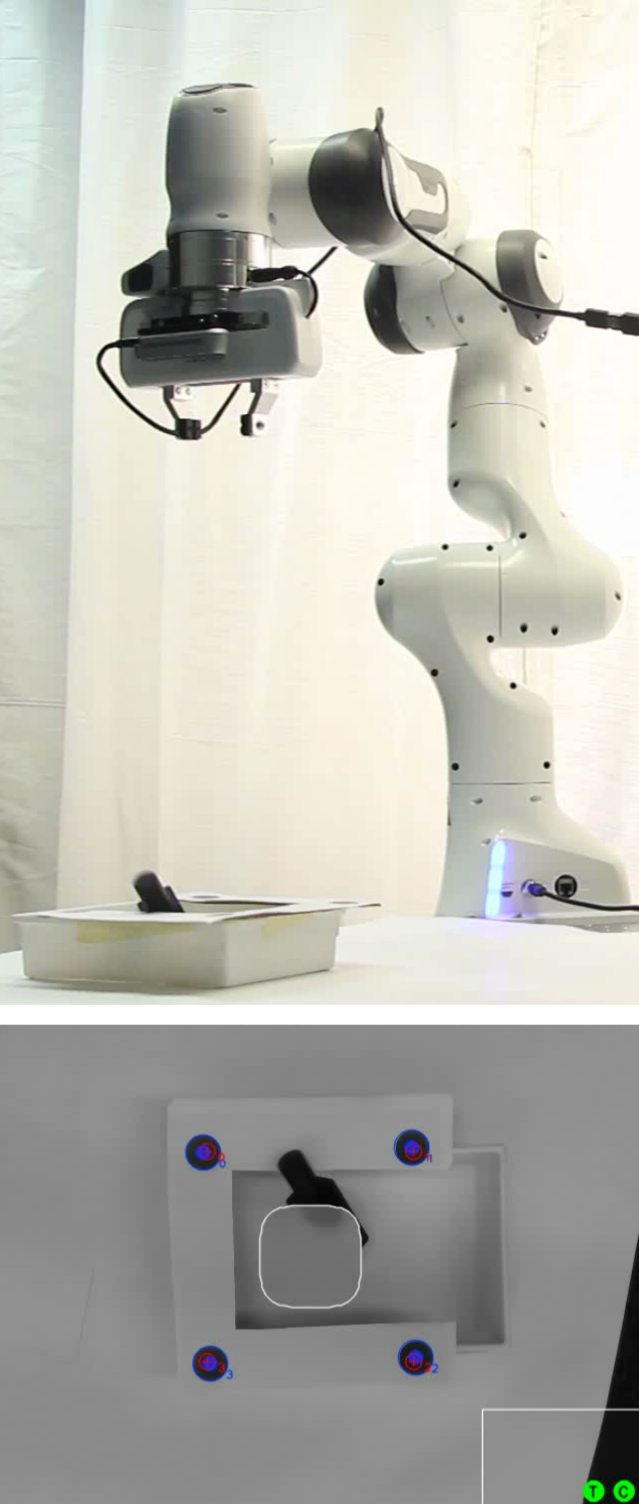}\label{fig:exp_c}}
\hfill
\end{center}
\caption{Robot experiment using the memory of motion: VPC is able to avoid the occlusion and achieve the desired task.}
\label{fig:exp}
\end{figure}

\begin{figure}[t!]
\begin{center}
\includegraphics[width=\columnwidth]{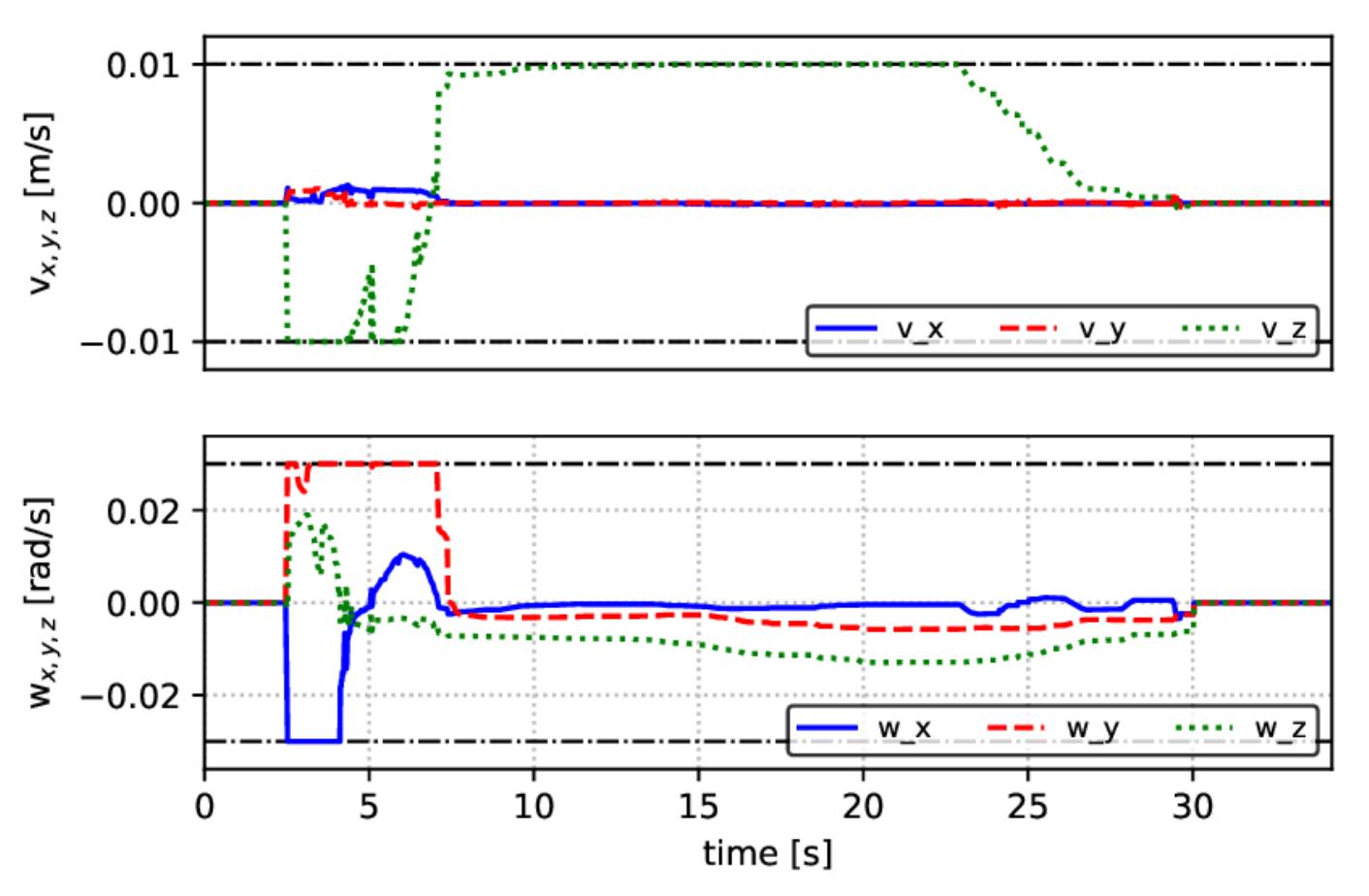}
\end{center}
\caption{Robot experiment using the memory of motion: velocity commands.}
\label{fig:vel_cmd_exp}
\end{figure}

For the on-line experiments, we set $N_p = 3$, the solver was given $10$ maximum iterations and $10^{-6}$ as optimality tolerance.
With this setting, and for some initial robot-box configuration, the previous iteration strategy was not capable to achieve the task, as shown in the snapshots of Fig.~\ref{fig:exp_fail}.
While moving the visual features (blue dots, see Fig.~\ref{fig:exp_fail_a}) towards the target (red circle), the features met the blurred spot (Fig.~\ref{fig:exp_fail_b}) causing the loss of a feature and the consequent failure of the task (Fig.~\ref{fig:exp_fail_c}).
The same experiment has been carried out with the GPR-based approach, see Fig.~\ref{fig:exp} where both robot and camera view are shown.
Starting from the same initial condition (Fig.~\ref{fig:exp_a}), at the proximity of the constraint (Fig.~\ref{fig:exp_b}), the memory provides proper way point (depicted as red crosses on the image plane) and warm-start which allow to successfully achieve the desired task (Fig.~\ref{fig:exp_c}). 
In Fig.~\ref{fig:vel_cmd_exp} are shown the velocity commands sent to the robot during the execution.
The experiments are shown in the accompanying video.

\section{CONCLUSION AND FUTURE WORK}\label{sec:conclusion}

In this paper, we addressed the problem of efficiently achieving visual predictive control tasks. 
Using a memory of motion, we could exploit previous solutions to better fulfill on-line tasks.
Furthermore, leveraging pre-computation contained in the memory, we could set a short VPC preview window without invalidating the results.
The algorithm performances rely on the pre-computed dataset; we plan to improve the quality of the memory using a global optimizer or a planner.
Furthermore, more sophisticated paradigm of active learning can be employed to build a minimal memory, containing less but more informative samples.
In the presented work, the memory is queried using k-NN and GPR. 
As shown with both simulations and experiments, these methods were able to outperform the standard VPC scheme.
However, we believe that the performance could be even improved by considering other kinds of regressors that can cope with multimodality, as done in~\cite{SantosoLembono:ral:2020}.
%
%
In the presented results we show that the use of a memory of motion helps also to keep the computation time limited.
However, more effort will be done in order to ensure full real-time performances.
Finally, further developments will be devoted to include the proposed scheme within the optimization framework of more complex systems such as humanoids.

\IEEEtriggeratref{15}



\bibliographystyle{IEEEtran}
\bibliography{warmVPC}

\end{document}